\documentclass[5p,times,twocolumn]{elsarticle}
\makeatletter
\def\ps@pprintTitle{%
	\let\@oddhead\@empty
	\let\@evenhead\@empty
	\def\@oddfoot{}%
	\let\@evenfoot\@oddfoot}
\makeatother

\usepackage{natbib}
\usepackage{multirow}
\usepackage{amssymb}
\usepackage{gensymb}
\usepackage{fontenc}
\usepackage{amsmath}
\usepackage{hyperref}
\usepackage{color}
\usepackage{graphicx,subfigure}

\usepackage{lineno}

\begin{document}

\begin{frontmatter}
	
	\title{Histological images segmentation of mucous glands}
	
	\author[cmc]{A.~Khvostikov\corref{cor1}}
	\ead{xubiker@gmail.com}
	
	\author[cmc]{A.~Krylov}
	\ead{kryl@cs.msu.ru}
	
	\author[med]{O.~Kharlova}
	\ead{olga.arsenteva@gmail.com}
	
	\author[med]{N.~Oleynikova}
	\ead{noleynikova@mc.msu.ru}
	
	\author[med]{I.~Mikhailov}
	\ead{imihailov@mc.msu.ru}

	\author[med]{P.~Malkov}
	\ead{pmalkov@mc.msu.ru}

	\cortext[cor1]{Corresponding author}
	
	\address[cmc]{Faculty of Computational Mathematics and Cybernetics, Lomonosov Moscow State University, Moscow, Russia}
	\address[med]{University Medical Center, Lomonosov Moscow State University, Moscow, Russia}
	
	\begin{abstract}
		Mucous glands lesions analysis and assessing of malignant potential of colon polyps are very important tasks of surgical pathology. However, differential diagnosis of colon polyps often seems impossible by classical methods and it is necessary to involve computer methods capable of assessing minimal differences to extend the capabilities of the classical pathology examination. Accurate segmentation of mucous glands from histology images is a crucial step to obtain reliable morphometric criteria for quantitative diagnostic methods. We review major trends in histological images segmentation and design a new convolutional neural network for mucous gland segmentation
	\end{abstract}
	
	
\end{frontmatter}

\section{Introduction}
\label{S:Introduction}

\begin{figure}[h]
  \centering
  \includegraphics[width=0.7\linewidth]{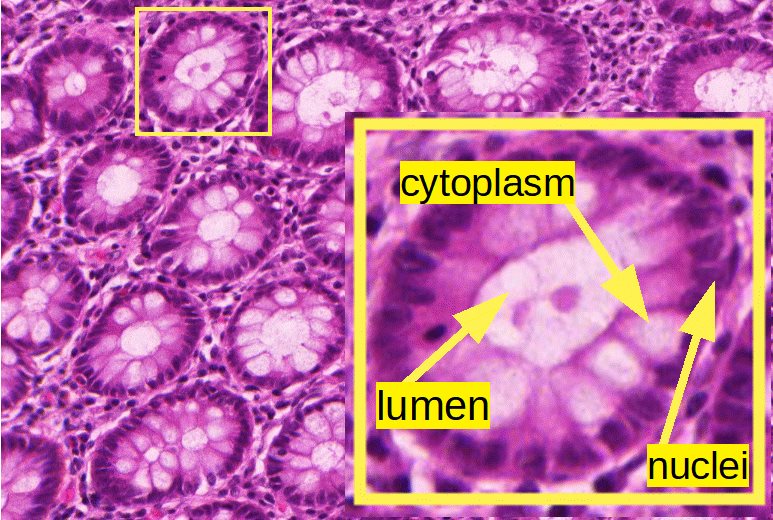}
  \caption{A histological image of colon mucous glands.}
  \label{fig:gland_parts}
\end{figure}

Mucous glands are important histological structures presented in certain organ systems as the main mechanism for secreting proteins and carbohydrates. Malignant tumors arising from glandular epithelium, also known as adenocarcinomas, are the most prevalent form of cancer. However for the mucous glands the sequence of precancerous processes leading to the development of a malignant tumor has been studied most well: metaplasia-dysplasia-cancer. The detection of metaplasia and dysplasia in the epithelium of the mucous glands may be an extremely difficult task for surgical pathology. Minimal differences between these processes are often not available for the classical pathology examination. An important practical task is to find analytical instrument for objective diagnosis.
Computational analysis is an additional analytical instrument for the segmentation of mucous glands on a microscopic images, shape analysis of the whole gland and it's lumen (Fig.\ref{fig:gland_parts}), calculating nuclear-cytoplasmic ratio in mucus-forming cells, and localization of the expression of certain immunohistochemical markers. All these parameters together can potentially significantly increase the accuracy of diagnosis of pathological processes associated with mucous glands in colon, stomach, prostate, breast, bronchus, etc.

Accurate segmentation of glands is often a crucial step to obtain reliable morphological statistics. Nonetheless, the task by nature is very challenging due to the great variation of glandular morphology in different histological grades. 

Today the standard dataset for the task of histological images segmentation is Warwick-QU image dataset \cite{Warwick-QU}, first used in \cite{stochastic_polygons2015}. The dataset provides 165 Hematoxylin and Eosin (H\&E) stained slides, consisting of a variety of histological grades. The dataset is provided together with ground truth annotations made by expert pathologists. Every image contains a number of colon glands, each of them consists of lumen, cytoplasm and nuclei (Fig.\ref{fig:gland_parts}). Warwick-QU dataset was also used in the Gland Segmentation in Colon Histology Images (GlaS) contest \cite{GLAS}.

The goal of this paper is to give a short overview of recent trends in histological images segmentation and propose a new algorithm for mucous glands segmentation. The target aim of our research is the computer-aided diagnostics of colon polyps – adenomas (ICD-O 8140/0) and serrated lesions (ICD-O 8213/0).

\section{Review of gland segmentation methods}
All major trends in gland segmentation can be roughly split into two groups (those which are based on the use of conventional image processing techniques and more modern methods based on convolutional neural networks), which are overviewed in Section \ref{sec:convantional_segm} and Section \ref{sec:networks_segm} accordingly. 

\subsection{Conventional methods of segmentation}
\label{sec:convantional_segm}
The first methods of image segmentation ever applied to histological images were methods based on K-means clustering, Chan-Vese algorithm, histogram analysis, edge detection, and watershed algorithm. Although these methods of segmentation are very simple, they are not effective enough, and can be used only for semiautomatic segmentation \cite{semiautomatic_segm}.

The next generation of algorithms was based on a more complex feature-based approaches and could perform automatic image segmentation in most cases. E.g. a region growing algorithm is presented in \cite{wu2005_region_growing}. The initial seeding regions are identified based on the lumen inside the glands by fitting with a large moving window. The seeding regions are then expanded by repetitive application of a morphological operation. False gland regions are removed based on either their excessive ages of active growth or inadequate thickness of dams formed by goblet cell nuclei outside the grown regions.

In \cite{object_graph} the tissue image is segmented using the definition of object-graphs instead of using the pixel-based approach. The image is decomposed into a set of circular objects, and depending on the organizational properties of these objects initial seeds are determined. After that the inner regions of glands are found by growing the seeds in accordance with nuclei location.

A more complex approach was proposed in \cite{stochastic_polygons2015}. Each gland in the image is treated as a polygon of random number of vertices, while the vertices represent approximate locations of epithelial nuclei. The problem of constructing such a graph is formulated as a Bayesian inference problem by defining a prior for spatial connectivity and arrangement of neighboring epithelial nuclei and a likelihood for the presence of a gland structure.

\subsection{Segmentation with neural networks}
\label{sec:networks_segm}

Recently, image segmentation methods using convolutional neural networks are gaining in popularity. Due to their good generalization capacity and versatility these methods demonstrate the state-of-the-art level of performance. Almost all CNN-based segmentation methods use the same idea of convolutional autoencoder (CAE) \cite{CAE}. With minor changes these CNN-based segmentation methods can be also applied to histological images.

The first approach of semantic image segmentation using CNN was Fully Convolutional Network (FCN) proposed by \citeauthor{FCNN} in \cite{FCNN}. The key ideas of the approach are to replace the fully-connected layers with convolutions, perform upsampling with transposed convolutions (or so called upconvolutions) and use skip connections. All in all FCN re-purposes imagenet pretrained networks for semantic segmentation problem.

Despite it's upconvolutional layers and a few shortcut connections FCN produces too coarse segmentation maps. SegNet \cite{segnet} proposes another way of upsampling. The decoder part of the SegNet uses pooling indices computed in the max-pooling step of the corresponding encoder part to perform non-linear upsampling. This leads to more accurate segmentation and makes SegNet more memory efficient than FCN.

The main contribution of U-Net \cite{unet} compared to other fully convolutional segmentation networks is that while upsampling and going deeper in the network the upsampled features are concatenated with the higher resolution features from down part for better localization. Furthermore, apart from proposing the new architecture and a special way of data augmentation U-Net was the first CNN-based method applied for biomedical image segmentation.

The main problem of the mentioned above approaches of segmentation is the impossibility of the algorithms to separate close or contiguous objects. Various ideas were proposed to solve this problem.

For example in \cite{Topology_2016} the authors introduced a new high-level loss function, that takes into account high level shape priors, such as smoothness and preservation of complex interactions between object regions. This loss was used in the FCN network trained on the the Warwick-QU dataset and demonstrated better performance compared to the FCN trained with conventional per-pixel loss.

In \cite{CNN_TV_2017} \citeauthor{CNN_TV_2017} trained two FCN networks on the Warwick-QU dataset. The first network performs a 4-class object segmentation (benign background, benign gland, malignant background and malignant gland), while the second one is trained to separate close glands. The CNN predictions are then regularized using weighted total variation to produce the final segmentation result. 

\citeauthor{DCAN_2017} in \cite{DCAN_2017} proposed a DCAN architecture with similar idea of object detection and separation, but unlike \cite{CNN_TV_2017} these two steps are performed simultaneously with one FCN-based network that has two outputs. First output predicts probabilities of gland object, while the second predicts the probability map of contours separating glands. The final segmentation masks are calculated using the threshold rule. To strengthen the training process DCAN uses 3 weighted auxiliary classifiers in the 3 deepest layers of the network.

The idea of splitting segmented glands got a further development in \cite{xu2017_gland_inst_segm}. The authors introduce a CNN with 3 pipelines: a FCN for the foreground segmentation, Faster R-CNN \cite{FasterRCNN} for the object detection and HED \cite{HED} for edge detection. All three pipelines a fused into one, and are followed with several convolution layers to predict the final instance segmentation map. This approach leads to the state-of-the-art level of segmentation accuracy.

\section{Proposed method}
\begin{figure}[h]
	\subfigure[Convolution block]{\label{fig:NN_blocks_conv}\includegraphics[width=0.4\linewidth]{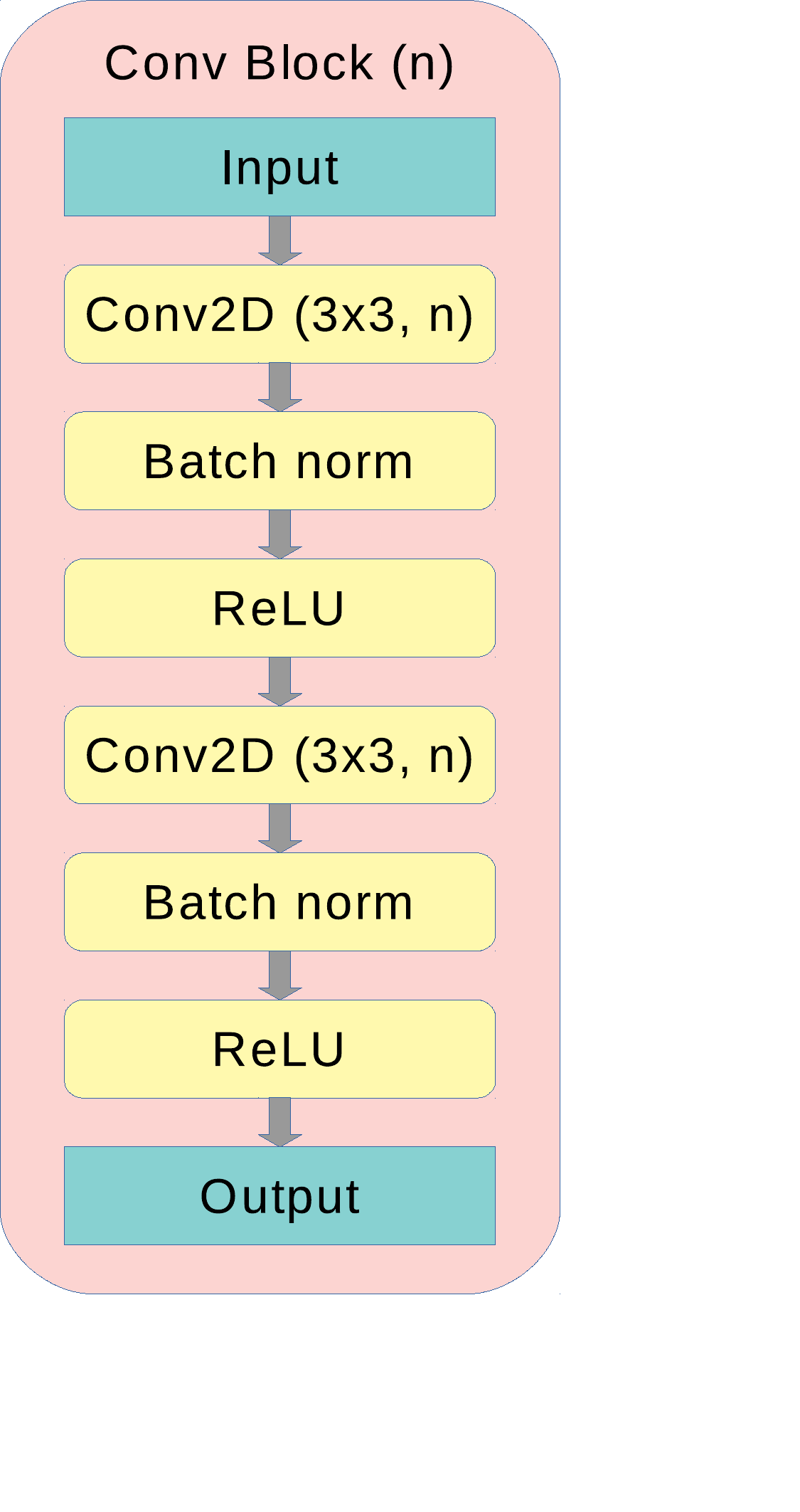}}
	\subfigure[Upconvolution block]{\label{fig:NN_blocks_upconv}\includegraphics[width=0.4\linewidth]{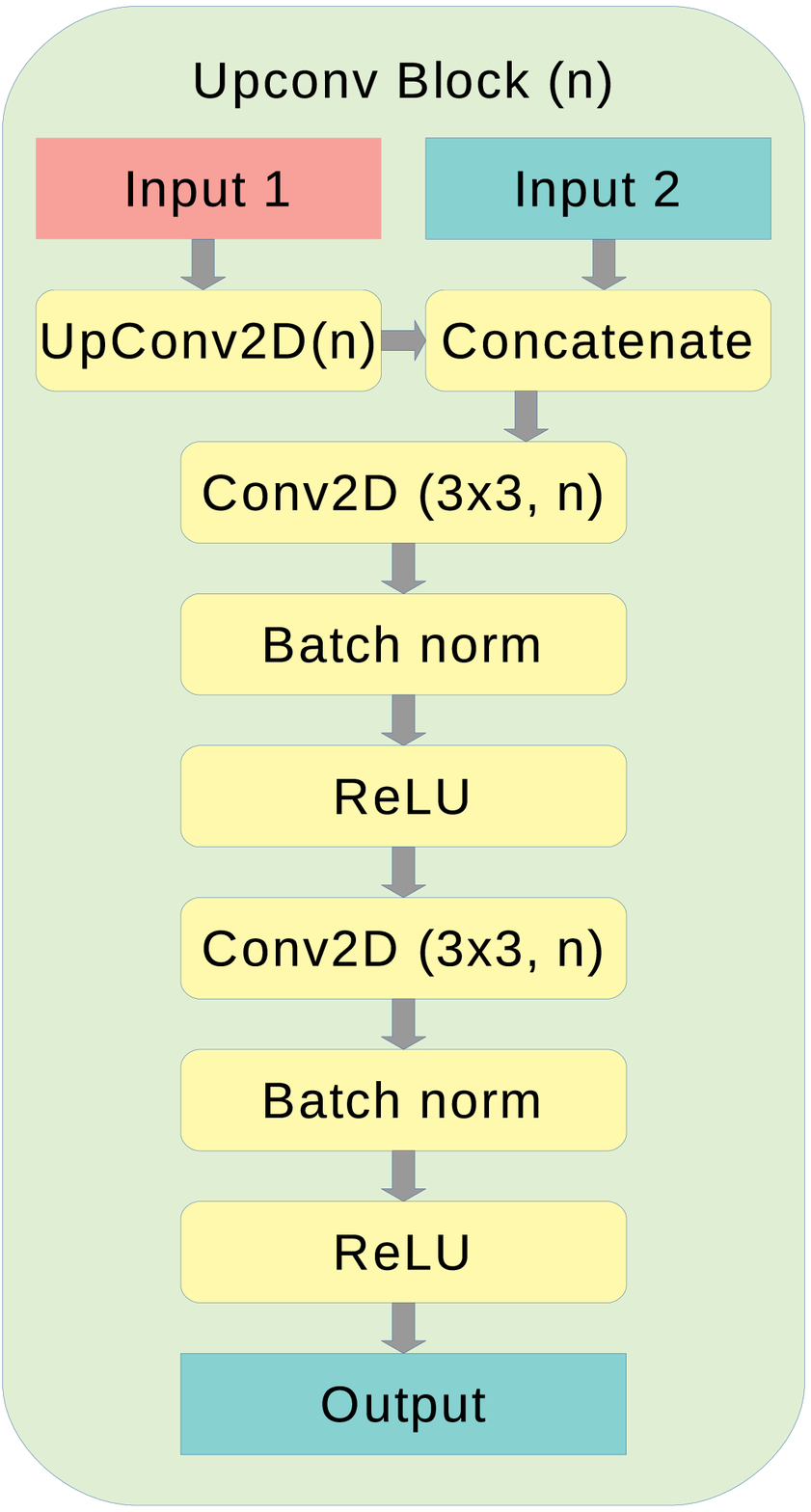}}
    \caption{Base building blocks of the proposed network.}
	\label{fig:NN_blocks}
\end{figure}

In this paper we consider the problem of histological mucous gland segmentation. For this problem we suggest a rather simple yet effective CNN architecture. The approach is based on the U-Net \cite{unet} with batch normalization \cite{batchnorm} layers added after each convolution. The main building blocks of the network are Conv block (Fig.\ref{fig:NN_blocks_conv}) and Upconv block (Fig.\ref{fig:NN_blocks_upconv}).

\begin{figure}[h]
  \centering
  \includegraphics[width=\linewidth]{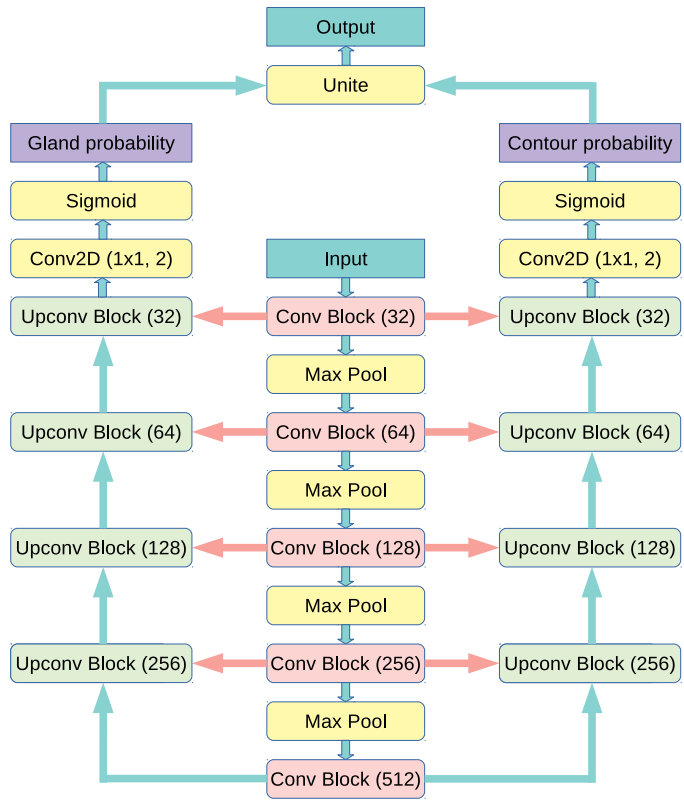}
  \caption{Proposed network architecture for mucous glands segmentation.}
  \label{fig:NN}
\end{figure}

Conv blocks form the contracting path of the network, while Upconv blocks form the expansive path. Here Conv2D (SxS, n) operation stands for the 2D convolution with $n$ filters of size $S\times S$, while UpConv2D (n) operation stands for the double upsampling of the feature map followed by 2D convolution with $n$ filters of size $2\times 2$. In contrast to the U-Net we use two different expansion paths in order to detect glands and separating contours. Each expansion paths ends with a $1\times 1$ 2D convolution, which reduces the number of features to 2 (background/foreground) and a sigmoid activation function to get the output probability map. Thus, the first expansion path outputs the probability of gland and the second - the probability of gland contour. Uniting these two probabilities by threshold we get the final result of segmentation (Fig.\ref{fig:NN}). The network is learned using RMSprop optimizer \cite{RMSprop}, dividing the gradient by a running average of its recent magnitude. We use a sum of a binary cross entropy value and a dice coefficient as the loss for both of the network outputs.

\section{Experiments and results}
\label{S:results}

The algorithm was evaluated on the Warwick-QU dataset \cite{Warwick-QU}, which consists of a wide range of histological grades from benign to malignant subjects. The dataset contains 85 train and 80 test images. We enlarged the dataset by 10 times using augmentation process with random shift, rotation, zoom and flip operations.


The proposed algorithm was implemented using open source neural network library Keras \cite{keras} with TensorFlow backend. The training was performed on a platform with Intel(R) Core(R) i7-6700HQ CPU and NVIDIA GeForce GTX 960M GPU.
Due to the relatively high resolution of the images from the used dataset, every processed image is first split into non-overlapping patches of $256 \times 256$ size, each of the patches is fed to the network and after that the output image is merged from the processed patches.

\begin{figure}[h]
  \centering
  \includegraphics[width=\linewidth]{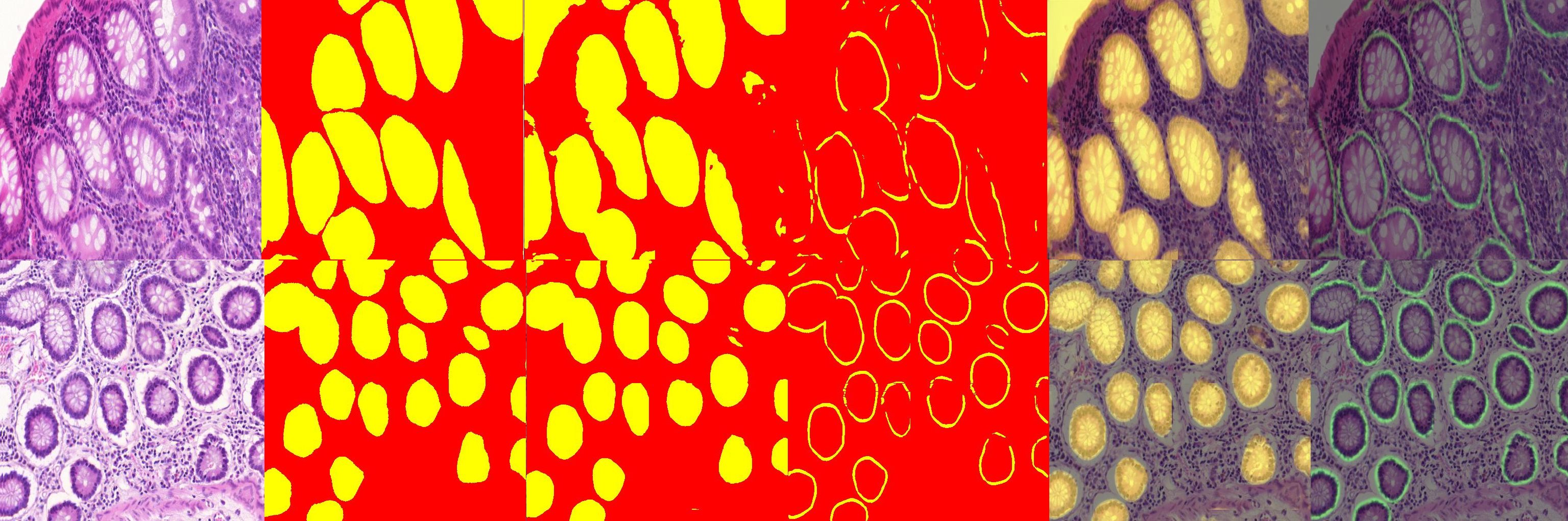}
  \caption{Segmentation results on the test images from dataset \cite{Warwick-QU}. The columns: source image, ground truth mask, network outputs (gland and contour detection), superpositions of source image and the outputs.}
  \label{fig:res}
\end{figure}

The segmentation results after training the proposed network for 12 epochs are shown in Fig.\ref{fig:res}. The evaluation on the test set demonstrates sufficiently good results in both gland and contour segmentation.

\begin{figure}[h]
  \centering
  \includegraphics[width=\linewidth]{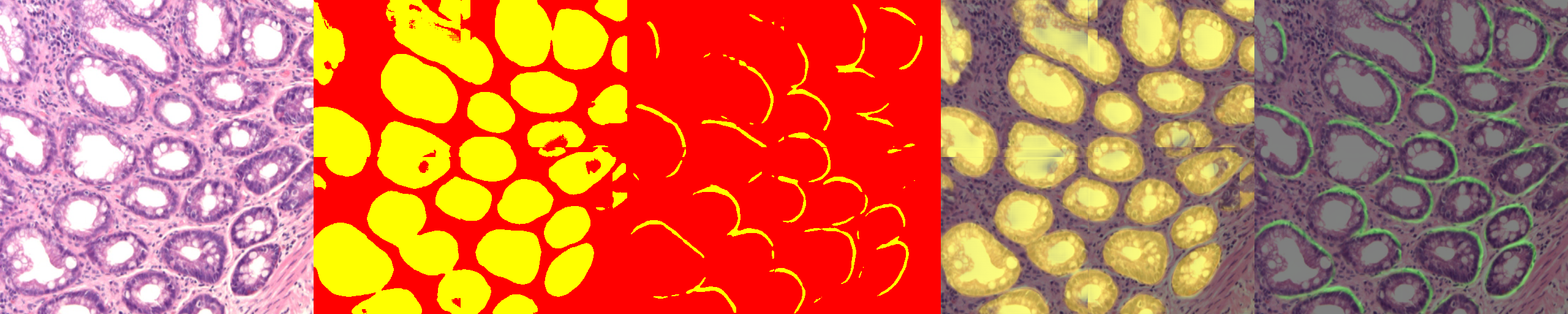}
  \caption{Segmentation results of an image fragment of mucous glands in hyperplastic colon polyp in real biopsy diagnostic material. The columns: source image; network outputs (gland and contour detection); superpositions of source image and the outputs.}
  \label{fig:res2}
\end{figure}

\section{Conclusion and further directions}

Further directions of the our work include performing a more complex segmentation to make also an inner-gland segmentation (detect nuclei, lumen and cytoplasm), which can be used for the ensuing analysis. In particular, analyzing the histological images of mucous glands helps to detect changes in its lumen shape (serration), in the nuclear-cytoplasmic ratio inside mucus-forming cells, and in the character of the expression of immunohistochemical markers \cite{kharlova2015_serrated}.

\section{Acknowledgements}

The work was supported by Russian Science Foundation grant 17-11-01279. In our study, the review of segmentation methods and designing the CNN were performed by the authors from the Faculty of Computational Mathematics and Cybernetics, the medical problem statements and experimental work - by the authors from the University Medical Center.

\bibliographystyle{plainnat}
\bibliography{biblio}

\end{document}